# A Multi-task Framework for Infrared Small Target Detection and Segmentation

Yuhang Chen, Liyuan Li, Xin Liu, Xiaofeng Su, and Fansheng Chen

*Abstract*—Due to the complicated background and noise of infrared images, infrared small target detection is one of the most difficult problems in the field of computer vision. In most existing studies, semantic segmentation methods are typically used to achieve better results. The centroid of each target is calculated from the segmentation map as the detection result. In contrast, we propose a novel end-to-end framework for infrared small target detection and segmentation in this paper. First, with the use of UNet as the backbone to maintain resolution and semantic information, our model can achieve a higher detection accuracy than other state-of-the-art methods by attaching a simple anchor-free head. Then, a pyramid pool module is used to further extract features and improve the precision of target segmentation. Next, we use semantic segmentation tasks that pay more attention to pixel-level features to assist in the training process of object detection, which increases the average precision and allows the model to detect some targets that were previously not detectable. Furthermore, we develop a multi-task framework for infrared small target detection and segmentation. Our multi-task learning model reduces complexity by nearly half and speeds up inference by nearly twice compared to the composite single-task model, while maintaining accuracy. The code and models are publicly available at https://github.com/Chenastron/MTUNet.

## I. Introduction

Infrared small target detection has significant applications in airspace surveillance, maritime surveillance, anti-missile and missile guidance, etc. Detection of multiple small targets on a single infrared image frame is a challenging task. First of all, the small target scale brings problems to the target detection task: the multi-layer convolution and down-sampling in the feature extraction backbone of common target detection networks can easily cause the loss of small target features [1]. Performance can be improved by fusing shallow features [2], but the contradiction between high-level semantics and high-resolution features cannot be solved fundamentally. In addition, target anchors and regression methods in common target detection heads don't take small targets much into account [3]. What's more, the number of negative samples in the images is too large due to the small and few targets, which account for most of the loss during training. Most negative samples are easily categorized, so the model cannot be optimized in the desired direction. These problems have led to an average precision (AP) gap of about 30 between large and small targets on the COCO test-dev benchmark now (RANK 1: DINO[4], $AP_L$:76.5, $AP_S$:46.7). Targets in infrared images cause even more difficult problems, such as a lack of shape and texture features, compared to targets in visible images. Overall, it is very difficult to detect infrared small targets by directly utilizing the network designed for normal objects [5–7].

Therefore, in most of the existing research on infrared small target detection, the detection task is regarded as a semantic segmentation task [8–13]. The reason why it works well may be that semantic segmentation actually classifies every pixel in the image, thus avoiding the influence of small target scale. However, most of those papers do not specify how to transform the result of semantic segmentation into the result of target detection, or just calculate the centroid of each target as the detection result rather than the typical bounding box. As far as we know, a robust, end-to-end, single frame based infrared small target detection algorithm has not been proposed.

Overall, the general object detection method cannot achieve accurate results when applied directly to infrared small targets. On the other hand, segmentation-based methods cannot transform the segmentation results into multi-target bounding boxes of appropriate size and cannot deal with overlapping targets. We need a robust and effective method to directly detect small targets, which is very necessary and valuable. Based on these motivations, we propose a multi-task Unet (MTUNet) framework for infrared small target detection and segmentation.

In our method, the target detection branch uses UNet as the target detection backbone [14] and the improved CenterNet as the target detection head [15], which greatly improves the

Manuscript submitted June 14, 2022; This work was supported by the Strategic Priority Research Program of the Chinese Academy of Sciences of the Shanghai Institute of Technical Physics, Chinese Academy of Sciences with the under Grant XDA19010102; National Science Foundation of China under Grant 61975222. (Corresponding author: Fansheng Chen).

Yuhang Chen, Liyuan Li and Xin Liu are with the Key Laboratory of Intelligent Infrared Perception, Shanghai Institute of Technical Physics, Chinese Academy of Sciences, Shanghai 200083, China, and also with the University of Chinese Academy of Sciences, Beijing 100049,China (e-mail: chenastron@163.com;liliyuan@mail.sitp.ac.cn;liuxin@mail.sitp.ac.cn ).

Xiaofeng Su is with Shanghai Institute of Technical Physics, Chinese Academy of Sciences, Shanghai 200083, China (e-mail: fishsu@mail.sitp.ac.cn).

Fansheng Chen* is with the Key Laboratory of Intelligent Infrared Perception, Shanghai Institute of Technical Physics, Chinese Academy of Sciences, Shanghai 200083, China, and also with the Hangzhou Institute for Advanced Study, University of Chinese Academy of Sciences, Hangzhou, China, 310024 (e-mail: cfs@mail.sitp.ac.cn).

performance of small target detection. The semantic segmentation branch also uses UNet as its backbone and uses the pyramid pooling module to collect multi-level feature information and fuse them [16], which improves the accuracy of segmentation predictions. These two processes are called single-task learning [17]. According to research, training multiple single tasks together via a multi-task network not only has a lightweight advantage but can also sometimes improve the performance of each task [18,19]. Therefore, we carry out multi-task learning for object detection that pays more attention to object-level features and semantic segmentation that pays more attention to pixel-level features. Results show that the performance of the two tasks is maintained at a similar level or even gets higher. In addition, compared with two separate single-task networks, MTUNet reduces the number of parameters and speeds up inference. The contributions of this paper are as follows:

1) To the best of our knowledge, we are the first to propose a general end-to-end infrared small target detection and segmentation method that uses UNet as the global backbone. We believe that the ability of UNet to fuse the low-level features and maintain the resolution of the feature map is very important for small object vision tasks. Then, just attaching a simple but effective detection or segmentation head can make the model achieve superior performance.
2) Using the improved CenterNetHead, the object detection is carried out directly on the full-resolution feature map, which greatly improves the target detection accuracy. We prove the advantage of the anchor-free method in small target detection.
3) In the part of semantic segmentation, the pyramid pool module is used to enhance multi-level feature information. The training method is specially designed for the situation where the small target accounts for only a few pixels in the image, which improves the precision of target segmentation.
4) Multi-task learning is carried out for infrared small target detection and segmentation, and the multi-task model can achieve improvement in all metrics. The experiments show that the shared backbone makes full use of the similar semantic features of the two tasks and can help the detection network deal with hard samples. Compared with the composite single-task model, MTUNet reduces the number of parameters and speeds up inference.

II. RELATED WORK

*A. Infrared Small Target Detection*

The traditional approach to infrared small target detection is directly establishing a model to measure the difference between an infrared small target and its surroundings. Some studies, such as [20,21], use multi-frame methods to detect infrared targets, but we only discuss single-frame methods in this paper. The single-frame detection problem has been modeled as target point detection under different assumptions, such as significant target points [22], sparse points in a low-rank background [23], and prominent points in a homogeneous background [24,25]. Accordingly, the infrared small target detection results can be obtained by significance detection, sparse low-rank matrix decomposition, or local contrast measurement. Finally, the small infrared target is segmented by using the threshold. Though these methods are computationally friendly and do not require training or learning, it is difficult to design features and superparameters artificially. In addition, they cannot adapt to changes in target shape and size, as well as complex backgrounds. Consequently, we have seen an increase in methods based on the deep neural network (DNN) model in recent years. These methods have been able to achieve much better performance than traditional methods because they can learn features from large amounts of data.

Existing DNN-based methods treat target detection as a semantic segmentation task. Dai et al. designed a semantic segmentation network using an asymmetric contextual module and introduced a dilated local contrast measure to improve their model [9,10]. Hou et al. proposed a network that combines manual features and convolution neural networks, which can adapt to different sizes of targets [11].

*B. Datasets for Infrared Small Targets*

Compared with visible image datasets, infrared image datasets have been trapped by the scarcity of public data for a long time. Most methods are trained and evaluated on their own private datasets. Wang et al. first established an open infrared small target data set, which contains 10k images. However, many targets in the dataset are too large and inaccurate, which affects the effect of training. [8]. Dai et al. published a dataset with high-quality semantic segmentation masks, but its target detection labels are inaccurate [9]. In order to perform the experiments on this dataset, we regenerate the target detection annotation in COCO format based on the segmentation annotation. Due to the size of the dataset, we divided it into 70% training dataset and 30% validation dataset.

Li et al. made a thermal infrared ship detection dataset using the SDGSAT-1 thermal imaging system [26]. However, thanks to the high resolution of SDGSAT thermal imager, the ship target has more pixels than the target studied in this paper. In [21], Du et al. used a dataset of infrared images for dim small aircraft target detection and tracking under ground/air background provided by [27], which is a video dataset containing 22 videos. However, each video has similar backgrounds and the same target, so there is high data redundancy between frames. We think it is not suitable for training a single-frame detection model for various targets.

*C. Multitask Learning*

The common single-task visual perception networks consist of various backbones and various heads. The widely used networks are UNet [14] and DeepLab [28] for semantic segmentation; Faster-RCNN [3] and YOLO [6] for target detection. In contrast, multi-task learning architectures apply

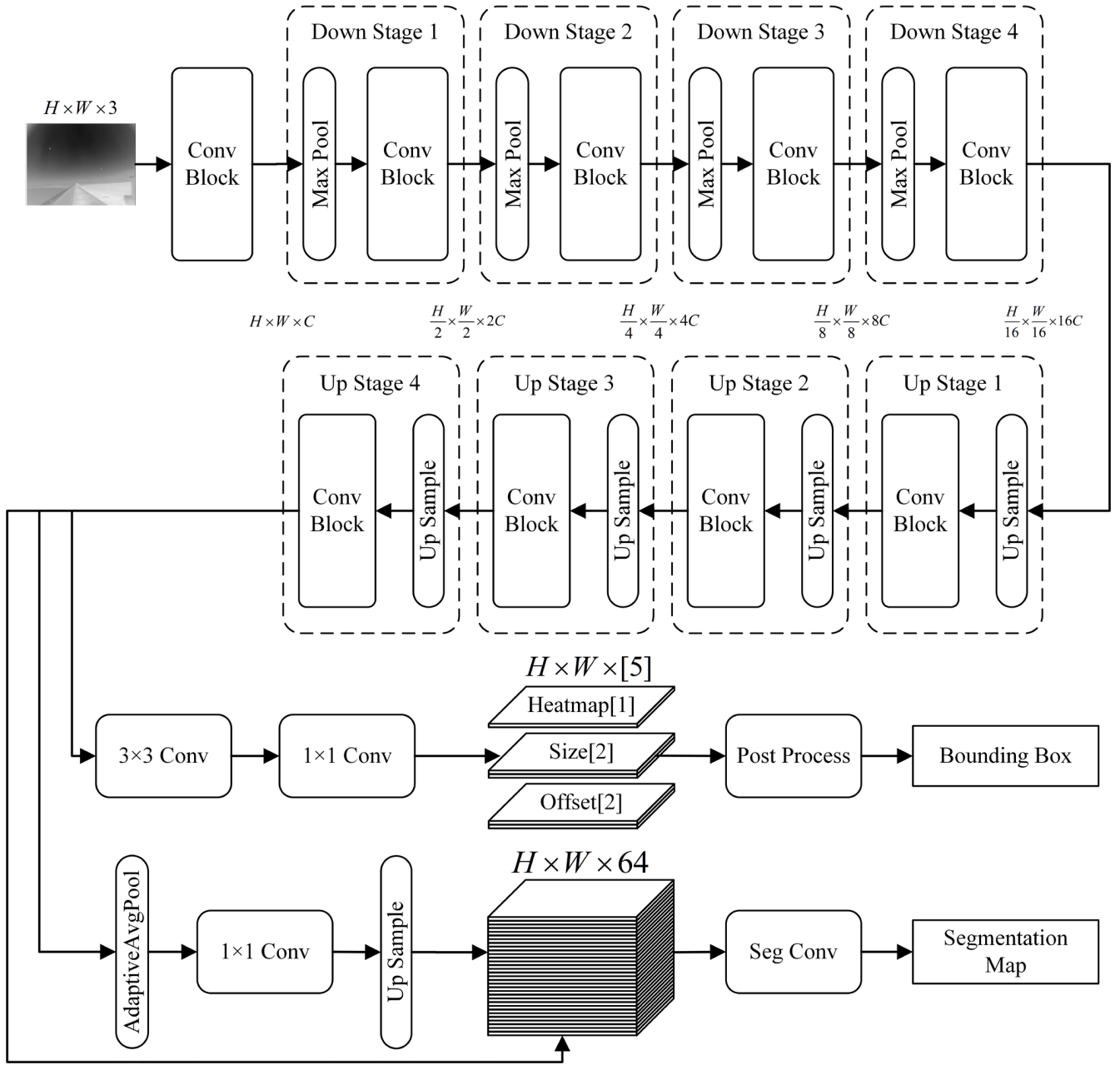

**Fig. 1.** Architecture of the proposed MultiTask-UNet.

different heads for different tasks on the same backbone module.

Multi-task learning can be seen as a transfer of inductive bias from the perspective of machine learning, which makes the model more inclined to certain assumptions. For example, CNN introduces two inductive biases, locality and translation equivariance, so that the model can obtain a lot of prior information. In multi-task learning, inductive biases are provided by other tasks. That makes the model more likely to learn features that can explain multiple tasks at the same time, which improves the generalization performance. Examples mainly related to visual perception tasks are Mask R-CNN [29], PanopticFCN [30].

By preserving and sharing the feature representations of different tasks, one task can help the training process of other tasks. A multi-task algorithm can combine the processes of single-task algorithms in many ways. One way is to use a backbone pre-trained on one task and then attach head modules for other tasks. In comparison with random initialization weights, it can not only speed up convergence but also improve performance. Another way is to directly construct a model containing a single backbone and multiple heads. By combining model structures and loss functions, multiple tasks can be trained simultaneously.

III. PROPOSED METHOD

In this section, we describe the overall architecture and optimization formulation of the proposed model.

The overall structure of the network is shown in **Fig. 1**, inputting the infrared image to be detected, then generating the semantic segmentation map and outputting the target detection bounding box. The scaled input image is processed by the UNet-type feature extraction backbone, and the generated feature map is sent to the semantic segmentation branch and the object detection branch, which finally outputs the semantic segmentation map of the target/background and the detection results of multiple targets.

In the existing research, UNet is not normally used as the backbone for object detection since it is not cost-effective. For objects of normal size, their semantic information can be retained in low-resolution feature maps after multi-layer convolution and down-sampling and can be recovered by up-sampling methods such as bilinear interpolation. Although this process is very efficient, it limits the model's ability to detect small targets. In contrast, UNet preserves the semantic information of small targets in the final feature map through learnable up-sampling and multi-scale feature fusion. The superiority of UNet in small target detection will be verified in comparative experiments later on. In addition, UNet performs well in the field of medical image segmentation. Medical images exhibit similar characteristics to infrared images in comparison with natural images. We selected UNet as the backbone for the thermal infrared domain because they both have simple image semantics, a small amount of data, and a need for high-resolution information.

It is difficult to detect small objects on a small scale because object detection focuses on object-level features. Therefore, our framework introduces semantic segmentation that focuses on pixel-level features for joint learning. Since semantic segmentation and target detection are related visual tasks, the semantic segmentation branch could assist in improving the performance of small target detection.

*A. Detection Branch*

In the common anchor-based target detection methods (such as Faster-RCNN), the anchor setting and regression of the bounding box do not take into account the problem of the target being too small, so we built our target detection head using CenterNet[15]. An anchor-free method like CenterNet can directly predict the coordinates, sizes, and offsets of key points with a simple and lightweight structure. To a certain extent, it also compensates for the influence of UNet on the number of model parameters and calculation amount.

The original CenterNet uses the up-sampling module with DCN to up-scale the output feature map of ResNet to 1/4 of the resolution of the original image before detection [31]. In our model, UNet is used as the backbone, and its output feature map is consistent with the resolution of the input image. Therefore, the up-sampling module is no longer required, and the feature map with the original image resolution is directly used for detection. The CenterNet head in our model actually acts at full resolution (Full-CenterNetHead).

Let's suppose the input feature map is $I \in R^{W \times H \times 64}$, and the detection branch produces three feature maps, namely keypoint heatmap, size prediction, and offset prediction.

**Keypoint heatmap:** When training, we need to produce a keypoint heatmap $\hat{Y} \in [0,1]^{W \times H \times C}$, where C represents the number of classes. In this task, C equals 1 because we only have one target class. $\hat{Y}_{x,y} = 1$ means a detected keypoint at $(x, y)$ while $\hat{Y}_{x,y} = 0$ means background. For each ground truth keypoint $p$, we splat it onto a heatmap $Y \in [0,1]^{W \times H \times 1}$ using a Gaussian kernel $Y_{xy} = exp\left(-\frac{(x-p_x)^2+(y-p_y)^2}{2\sigma_p^2}\right)$. $\sigma_p$ is an object size-adaptive standard deviation [32]. If two Gaussian kernels overlap, we take the element-wise maximum [33]. The loss between predict heatmap and ground truth heatmap is a pixel-wise focal loss[7]:

$$L_k = \frac{-1}{N} \sum_{xy} \begin{cases} \left(1-\hat{Y}_{xy}\right)^\alpha \log\left(\hat{Y}_{xy}\right) & \text{if } Y_{xy} = 1 \\ \left(1-Y_{xy}\right)^\beta \left(\hat{Y}_{xy}\right)^\alpha \log\left(1-\hat{Y}_{xy}\right) & \text{otherwise} \end{cases} \quad (1)$$

where $\alpha$ and $\beta$ are hyperparameters of focal loss[7], N is the number of keypoints. In our experiment, $\alpha$ and $\beta$ are set to 2 and 4, following[32].

**Size prediction:** Assuming that $(x_1^{(k)}, y_1^{(k)}, x_2^{(k)}, y_2^{(k)})$ is the bounding box of target $k$. Its center point lies at $p_k = \left(\frac{x_1^{(k)}+x_2^{(k)}}{2}, \frac{y_1^{(k)}+y_2^{(k)}}{2}\right)$. We regress the size of each object to $s_k = \left(x_2^{(k)} - x_1^{(k)}, y_2^{(k)} - y_1^{(k)}\right)$, which is computed before training process. We use L1 function to calculate loss of the prediction $\hat{S} \in \mathcal{R}^{W \times H \times 2}$ only at the center position of target:

$$L_{size} = \frac{1}{N} \sum_{k=1}^{N} \left|\hat{S}p_k - s_k\right| \quad (2)$$

**Offset prediction:** In the original CenterNet, due to the downsampling by a factor $R = 4$, the output feature map will introduce accuracy errors when remapped to the original image size. Thus, for each keypoint, an additional local offset is used to compensate for the error. In our model, due to the fact that the size of the feature map is the same as the input image, there is no such precision error. This means that we do not have to predict offsets. Because offset prediction just has a minimal effect on the inferencing speed, you can also keep it like CenterNet does[15]. Offset prediction does not affect the accuracy of object detection in our model.

The overall loss function is the sum of the loss of the keypoint heatmap loss $L_k$ and size loss $L_{size}$:

$$L_{det} = L_k + \lambda_{size} L_{size} \quad (3)$$

where $\lambda_{size} = 0.1$, following [15].

All outputs use the same feature map produced by UNet backbone. The features are passed through separate processes of $3 \times 3$ convolution, ReLU, and another $1 \times 1$ convolution. **Fig. 2** shows an overview of the detection branch final output.

In the postprocessing, we extract the locations of peaks in the heatmap as keypoint locations. Peak values are used to measure detection confidence, and bounding boxes are generated based on keypoint coordinates and size predictions.

*B. Segmentation Branch*

This branch uses the semantic segmentation method to predict the target and background from feature maps with the same

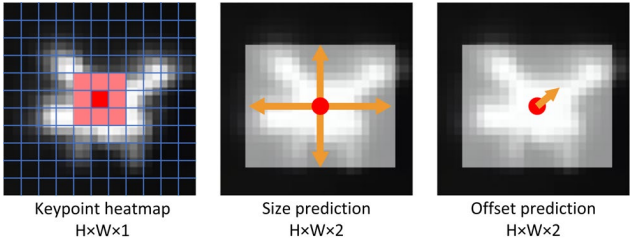

**Fig. 2.** An overview of the detection branch output. All predictions are produced from the same input feature map. Offset prediction does not affect the result and can be discarded.

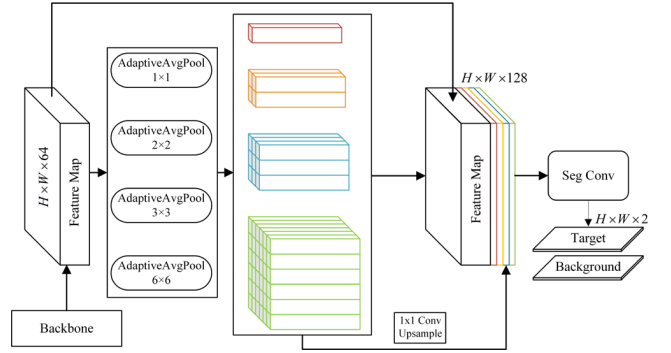

**Fig. 3.** Illustration of the process of the segmentation branch.

solution as input images.

In semantic segmentation, the background is often ignored and not included in training. This will cause difficulties in the infrared small target segmentation task since the model will learn the background as the target class. Because the proportion of target pixels in our image is very small, the model cannot be optimized in the desired direction. If most of the areas in the image are ignored, resulting in a high proportion of misclassified background pixels, such a model will be useless. In order to obtain the ideal training result, the background is seen as a category that can be learned in our model.

The input feature map is decoded by the pyramid pooling module to produce four feature bins of $1 \times 1$, $2 \times 2$, $3 \times 3$, $6 \times 6$ like [16], and then $1 \times 1$ convolution is used to reduce the number of channels from 64 to 16. The feature maps of these four scales are upsampled by bilinear interpolation, and then a concatenate operation is carried out on the channel dimension with the input feature map. Then pooling and convolution operations are performed again, and the last activation layer selects the sigmoid layer to generate the final segmentation prediction. **Fig. 3** shows the process of the segmentation branch.

We use dice loss [34] to train this branch, optimizing a weighted smooth dice loss $L_{seg}$ between prediction segmentation map and ground truth segmentation map:

$$Dice_{smooth} = \frac{2 \sum_{pixels} y_{true} \, y_{pred} + smooth}{\sum_{pixels} \left( y_{true}^2 + y_{pred}^2 \right) + smooth} \quad (4)$$

$$L_{seg} = \sum_{cls} w_{cls} \left( 1 - Dice_{smooth} \right) \quad (5)$$

where $y_{true}$ and $y_{pred}$ represent the corresponding pixel values from the ground truth mask and segmentation prediction. We use $smooth = 1$ for computational stability. We calculate $Dice_{smooth}$ on the target and background classes, respectively. The weights $w_{cls}$ of class target and class background are 10 and 0.1.

*C. Multi-task Loss Aggregation*

Like other multi-task joint learning studies, we can train these two visual tasks at the same time. The object detection branch uses $L_{det}$ for training, and the semantic segmentation branch uses $L_{seg}$ for training. The total loss of MTUNet is formulated as:

$$L_{all} = \lambda_{det} L_{det} + \lambda_{seg} L_{seg} \quad (6)$$

according to loss values in the beginning stage of training, we set $\lambda_{det} = 3$ and $\lambda_{seg} = 1$ to balance the effect of individual heads on the total loss. Performance metrics could be affected by different weight balances.

IV. EXPERIMENT AND ANALYSIS

In this section, we will introduce the implementation details and evaluation metrics, and compare our model with other methods on the same dataset. Experiments on semantic segmentation, object detection, and multi-task learning are carried out. To verify the effectiveness of our proposed model, we perform comparative experiments and ablation experiments.

*A. Implementation Details*

We used the public SIRST[9] dataset for experiments. This dataset has accurate semantic segmentation annotation, but its object detection annotation is very rough. Therefore, we generate the COCO format target detection annotation based on the segmentation annotation. Considering the quantity of data in this dataset, we divided it into a 70% training dataset and a 30% verification dataset.

We use random cropping and random flipping as data enhancement, and resize the input image to $320 \times 320$. We employ an SGD optimizer for 200 epochs with an exponential decay learning rate scheduler. The initial learning rate is 0.001 and a weight decay of 0.05 is used. The batch size is set to 4.

*B. Evaluation Metrics*

We use mIoU as the evaluation metric for the semantic segmentation task and AP as the evaluation metric for the target detection task, following protocols widely adopted in computer vision research. It will establish the consistency between infrared image and visible image tasks and contribute to the development of infrared vision research.

For the semantic segmentation task, the usual way is to generate predicted segmentation maps for each category contained in the dataset. Use IoU to compare predicted segmentation maps with ground truth masks, and mIoU to analyze the average performance of the model on all classes. There is only one target class in the SIRST dataset, so we can compare the target class segmentation results with other studies. Furthermore, as mentioned before, our model treats the target and background as separate categories, so we can also calculate the IoU of the background class and the mIoU of all classes.

TABLE I
SINGLE-TASK SEGMENTATION PERFORMANCE OF DIFFERENT METHODS

| Method | Target IoU | Background IoU | mIoU |
|---|---|---|---|
| IPI [35] | 25.67 | - | - |
| MDvsFA-cGAN [8] | 61.03 | - | - |
| ACM [9] | 71.65 | - | - |
| ALCNet [10] | 74.70 | - | - |
| DNANet-ResNet34 [13] | 76.98 | - | - |
| **Ours** | **78.94** | **99.98** | **89.46** |

We use COCO evaluation metrics to evaluate the performance of the target detection part. As with most detection tasks, we report average precision over all IoU thresholds (AP), AP at IoU thresholds 0.5 ($AP_{50}$), and 0.75 ($AP_{75}$). Considering the size of small targets, a deviation of a few pixels may cause large fluctuations in IoU. An IoU of 50% between the predicted bounding box and the ground truth label indicates a reasonable enough prediction, so $AP_{50}$ might be a more useful and fair metric.

*C. Experiment of Semantic Segmentation*

In the first part of experiments, we compare the semantic segmentation results of our network with other advanced methods, including IPI [35], which performs best in traditional methods, and data-driven methods such as MDvsFA-cGAN [8], ACM [9], ALCNet [10], and DNANet [13]. We evaluate these methods on our own dataset partition, and the results are compared as shown in **Table I**. It can be seen that the DNN-based method is clearly superior to the traditional method and our single-task semantic segmentation part has made a significant improvement over other DNN methods, reaching the highest target IoU of 78.94. In addition, the background IoU is 99.98 and mIoU is 89.46, which are not evaluated in other methods.

**Fig. 4** shows representative examples of semantic segmentation results. There is a lot of noise similar to the target in image (3). ALCNet cannot make the correct prediction. However, our method can clearly distinguish the target from the background. Our model also performs well with multiple targets in image (5). In contrast, ALCNet cannot distinguish all targets. ALCNet also have some false alarms in some cases with a high noise level or a low SCR. The performance of DNANet is similar to ours, but its structure is more complex, and its segmentation result is less accurate than ours. Our method has the best segmentation results in all kind of scenes.

*D. Experiment of Object Detection*

In the second part, experiments demonstrate the effectiveness of UNet as a target detection model backbone and the performance of the improved CenterNet in small target detection. Our model is compared with other architectures in two aspects: backbone and detection head. **Table II** shows the results.

**Comparison on backbones:** ResNet18, ResNet-34, and ResNet-50 are chosen as backbones for comparison. ResNet101 and ResNet152 are not used because they have many more parameters and computations than UNet.

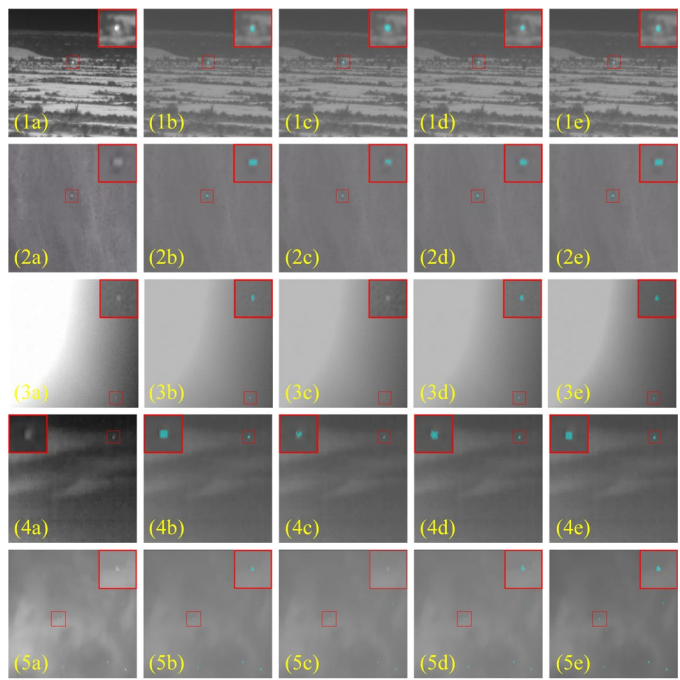

**Fig. 4.** Segmentation results of images with (1) complex background, (2) low SCR, (3) high noise level, (4) dim target, (5) multiple targets. The target area is painted in blue and the background area is painted in gray. (a) Input images. (b) GT labels. (c) ALCNet. (d) DNANet. (e) Ours.

It can be seen that UNet is +3.6 $AP_{50}$ higher (95.6 vs. 92.0) than ResNet-50, which proves that UNet is very suitable for infrared small target detection. It should be noted that ResNet is pre-trained on ImageNet. The absence of pre-training will result in significant performance degradation or even make the model impossible to optimize. UNet is trained from scratch and achieves better results, showing that UNet is more suitable for data-scarce infrared vision tasks than ResNet.

**Comparison on detection heads:** We compare the commonly used RetinaNet [7] with ours. RetinaNet uses Focal Loss to refine the model by focusing on hard samples. This problem is also one of the difficulties in infrared small target detection. RetinaNet surpassed the accuracy of the two-stage network and the speed of other one-stage networks. For the first time, the one-stage network had completely surpassed the two-stage network.

In the original CenterNet, ResNet is used to downscale the input image by a factor of 32, and CenterNetHead is used to make predictions on the upsampled feature map. We apply a full-scale strategy to CenterNetHead by performing two additional upsampling operations and then making predictions on a feature map of the same size as the original image. We call it Full-CenterNetHead, and the original one is called D4-CenterNetHead. D4 means downsampling by a factor of 4.

RetinaNet is much more complex than CenterNet, but D4-CenterNetHead outperforms RetinaHead by +2.0 $AP_{50}$ (89.3 vs. 87.3) using ResNet-50 backbone. Results show the superiority of anchor-free methods like CenterNet in detecting small targets. Full-CenterNetHead outperforms D4-CenterNetHead by +2.7 $AP_{50}$ (92.0 vs. 89.3), which shows effectiveness of full-scale strategy.

TABLE II
SINGLE-TASK DETECTION PERFORMANCE OF DIFFERENT METHODS

| Method | Backbone | $AP_{50}$ | $AP_{75}$ | AP |
|---|---|---|---|---|
| CenterNet | ResNet-18 | 72.40 | 4.90 | 22.60 |
| CenterNet | ResNet-34 | 74.00 | 6.80 | 23.40 |
| CenterNet | ResNet-50 | 89.30 | 32.40 | 38.50 |
| RetinaNet | ResNet-50 | 87.30 | 25.30 | 38.40 |
| CenterNet† | ResNet-50 | 92.00 | 27.70 | 40.40 |
| **CenterNet†** | **UNet** | **95.60** | **52.40** | **53.60** |

† indicates that a full-scale strategy is applied to the CenterNetHead. Full-CenterNetHead makes predictions on feature maps of the same size as the original images.

TABLE III
SINGLE-TASK VS. MULTI-TASK RESULTS

| Network | Target IoU | mIoU | $AP_{50}$ | $AP_{75}$ | AP |
|---|---|---|---|---|---|
| Only Seg | 78.94 | 89.46 | - | - | - |
| Only Det | - | - | 95.60 | **52.40** | **53.60** |
| Only Seg (Det Pretrained) | **79.72** | **89.85** | - | - | - |
| Only Det (Seg Pretrained) | - | - | **98.00** | 45.50 | 49.90 |
| MTUNet (Seg + Det) | 78.79↓ | 89.40↓ | 97.50↑ | 50.80↓ | 52.10↓ |

↑ means outperforming than the single-task model. ↓ means the opposite.

**Comparison on $AP_{75}$ and AP:** Our model achieves the highest $AP_{50}$ of 95.60, outperforming all other architectures. It is worth noting that compared with the improvement of $AP_{50}$ (+3.6), the network using UNet as the backbone has a greater improvement in $AP_{75}$ (+24.7) and AP (+13.2). This shows that the model can predict a more accurate and suitable bounding box, further proving the effectiveness of Unet backbone.

*E. Experiment of Multi-task Learning*

In the third part, we design and perform the multi-task experiments, combining the object detection task that prioritizes object-level features with the semantic segmentation task that emphasizes pixel-level features. Multi-task learning is done in two modes. **Table III** shows the comparisons of single-task and multi-task learning.

**Pre-train mode:** Pre-train the backbone for a detection or segmentation task and then use it to train another. Single-task models with the backbone pre-trained outperform the counterparts trained from scratch by +0.78 target IoU, +0.39 mIoU, and +2.4 $AP_{50}$. The results indicate that the semantic features learned by one task can benefit the training process of another task.

**Parallel mode:** We also train segmentation and detection tasks in parallel. MTUNet brings 1.9 $AP_{50}$ gains over single-task training, with a similar performance on semantic segmentation. The results indicate that the shared backbone exploits the similar semantic features of the two tasks. Compared with the Only Det (Seg Pretrained) model, the accuracy is slightly lower (97.5 vs. 98.0). Since MTUNet can also provide segmentation predictions, you can choose what is most appropriate for your needs.

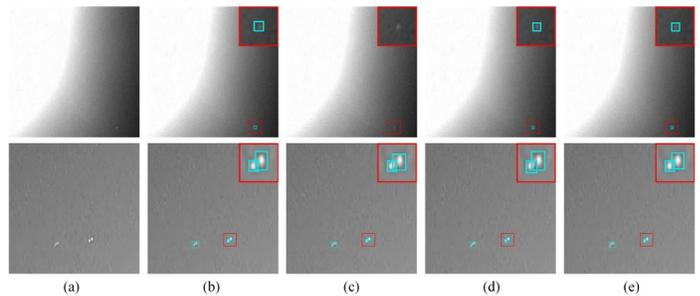

Fig. 5: A qualitative analysis of multi-task learning. (a) Input images. (b) GT labels. (c) Only Det, (d) Only Det (Seg Pretrained), and (e) MTUNet are visualized on the hard samples.

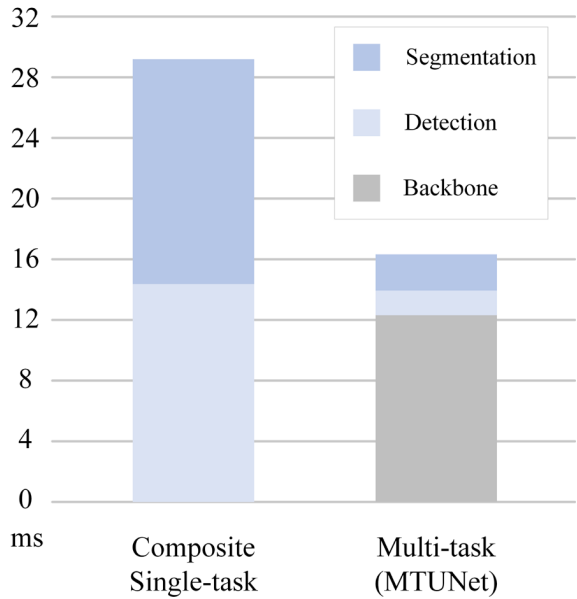

**Fig. 6** Inference time with Unet backbone. Benchmark with 2000 images and take the average. A composite single-task model takes 29.4ms to test an image. In contrast, MTUNet achieves a faster inference time of 16.8ms.

By comparing the visualization results of the single-task and multi-task detection model, you can also find that the semantic segmentation task is very helpful for the target detection task. In **Fig. 5**, the infrared image above contains a single target and a lot of noise similar to the target. The Only Det model is unable to detect the target effectively, but the multi-task model can correctly distinguish the hard sample. There are two very close targets in the image below. In the segmentation map, the two targets are connected. Our detection model can detect these two overlapping targets separately and a multi-task learning model can obtain more appropriate bounding boxes.

**Fig. 6** shows the inference time of single-task and multi-task models. Networks are evaluated on a NVIDIA GeForce RTX 3070 GPU. A resolution of 320×320 is chosen. MTUNet can train multiple tasks simultaneously and shorten the training time. Since the head module of our model only occupies a small part of the inference time, the inference time (16.8ms) is only slightly longer than that of the single-task semantic segmentation (14.9ms) and target detection (14.5ms). Our single-task model and multi-task model can both satisfy the requirement for real-time inference.

TABLE IV
COMPARISON OF DIFFERENT MODELS

| Models | Backbone | Head | params | FLOPs | FPS | $AP_{50}$ |
|---|---|---|---|---|---|---|
| RetinaNet | ResNet50 | FPN+RetinaHead | 36.10 M | 20.44 G | 46.4 | 87.3 |
| CenterNet | ResNet50 | D4-CenterNetHead | 46.85 M | **16.64 G** | 63.3 | 89.3 |
| CenterNet | ResNet50 | Full-CenterNetHead | 46.94 M | 29.79G | 54.6 | 92.0 |
| Only Det | UNet | Full-CenterNetHead | **29.06 M** | 86.85G | **69.0** | 98.0 |
| Only Seg | UNet | PSPHead | 29.05 M | 77.38G | 67.3 | - |
| MTUNet | UNet | Multi-head | 29.06 M | 88.78G | 59.5 | 97.5 |

*F. Network Structure Analysis*

In this section, we analyze the network structure in more detail. **Table IV** shows the representative target detection methods discussed in this paper, along with the components they use.

We can see that our model achieves the highest inference throughput of 69.0 images/s and the highest $AP_{50}$ of 95.60 with the fewest parameters. Although the UNet module produces a large amount of computation, it does not affect the inference speed probably because of its simple structure and high parallelism. Anchor-based methods such as RetinaNet are more complex, time consuming, have lower throughput, and are less accurate than anchor-free methods such as CenterNet. Compared with Full-CenterNetHead, the original D4-CenterNetHead predicts on the feature map with 1/4 of the resolution of the original image. Although it reduces the computation, the $AP_{50}$ is reduced by 2.7 (89.3 vs. 92.0). Our model gives a better speed/accuracy trade-off.

Thus, we consider UNet+CenterNet to be an effective paradigm for detecting small targets. Using an efficient backbone for generating feature maps of the same size as the input image and a simple detection head with a full-scale strategy, both speed and accuracy can be improved.

Moreover, since the UNet module accounts for more than 99% of the parameters, the number of parameters and calculations in MTUNet are nearly identical to those in the Only Det model. MTUNet can achieve 97.50 $AP_{50}$ and can detect some difficult samples that couldn't be detected before. Its shared backbone can reduce the model's storage requirements. If you need both segmentation and detection predictions, MTUNet is almost half the size of two single-task models, making it more suitable for deployment on edge or embedded devices.

## V. CONCLUSION

In this paper, we propose a multi-task learning model for infrared small target segmentation and location. Firstly, the potential of UNet as the backbone of the infrared image visual model is well explored in the single task, and the results demonstrate the advantage of the anchor-free method for detecting small targets. Then we apply the multi-task learning method and achieve similar or even better results on different tasks. Our model is suitable for detecting small objects in infrared images with complex backgrounds. It can detect targets with pixel-level accuracy, provide accurate semantic segmentation masks, and satisfy the real-time requirements. Compared with the composite single-task network, MTUNet reduces about half of the parameters and computations, so it is more suitable for deployment on the edge or embedded devices.

In addition, the multiple branches of our model can be trained simultaneously in an end-to-end manner, and the model can add new branches when given additional tasks and corresponding data annotations. MTUNet should be able to learn similar and general features from other tasks, such as super-resolution and deblurring, thus improving model accuracy and generalization performance. Our work is helpful to promote the research of infrared small target detection and can even be extended to other fields, such as small object detection in natural images.

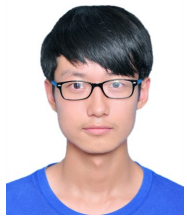

**Yuhang Chen** received the B.S. degree in computer science and technology from Dalian University of Technology, Dalian, China, in 2020. He is currently pursuing the M.A. degree in signal and information processing at Shanghai Institute of Technical Physics of the Chinese Academy of Sciences, Shanghai, China.

His current research interests include infrared small target detection through deep learning.

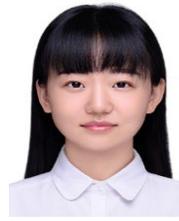

**Liyuan Li** received the B.S. degree in opto-electronics information science and engineering from Dalian University of Technology, Dalian, China, in 2018. She is currently pursuing the Ph.D. degree in physical electronics at Shanghai Institute of Technical Physics of the Chinese Academy of Sciences, Shanghai, China.

Her current research interests include dim and small targets detection of IR through machine learning.

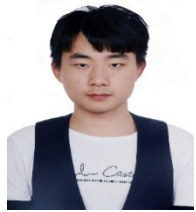

**Xin Liu** received the B.S. degree in Wuhan University of Science and Technology, Wuhan, China, in 2017. He is currently pursuing the Ph.D. degree in electronic circuit and system at Shanghai Institute of Technical Physics of the Chinese Academy of Sciences, University of Chinese Academy of Sciences, Beijing, China.

His current research interests include image processing, machine learning, and high-performance computing.

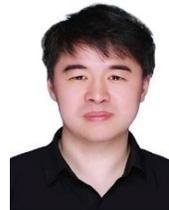

**Fansheng Chen** received the B.S degree in optoelectronic information engineering and Ph.D. degree in physical electronics from Shandong University, Jinan, China, in 2002 and Shanghai Institute of Technical Physics of the Chinese Academy of Sciences, Shanghai, China, in 2007, respectively.

Since 2013, he has been a Professor with the Shanghai Institute of Technical Physics of the Chinese Academy of Sciences. His research interests include the design of spatial high resolution remote sensing and detection payloads, high-speed and low noise information acquisition technology, and infrared dim small target detection technology. Meanwhile, he has been committed to the research and development of the space infrared staring detection instruments, the high spatial and temporal resolution photoelectric payloads, and the application of infrared multi-spectral information acquisition technology in artificial intelligence, target recognition and other relative aspects.